\documentclass[letterpaper]{article} % DO NOT CHANGE THIS
\usepackage{aaai2026}  % DO NOT CHANGE THIS
\usepackage{times}  % DO NOT CHANGE THIS
\usepackage{helvet}  % DO NOT CHANGE THIS
\usepackage{courier}  % DO NOT CHANGE THIS
\usepackage[hyphens]{url}  % DO NOT CHANGE THIS
\usepackage{graphicx} % DO NOT CHANGE THIS
\usepackage{natbib}  % DO NOT CHANGE THIS AND DO NOT ADD ANY OPTIONS TO IT
\usepackage{caption} % DO NOT CHANGE THIS AND DO NOT ADD ANY OPTIONS TO IT
\usepackage{algorithm}
\usepackage{algorithmic}

\usepackage{newfloat}
\usepackage{listings}
\usepackage{amsthm}       % 用于定义定理、引理等
\usepackage{bbding}       % 用于显示特殊符号
\usepackage{microtype}    % 自动优化排版，通常放在最后导入
\usepackage{amssymb}
\usepackage{amsmath}
\usepackage{booktabs}
\usepackage{multirow}
\DeclareCaptionStyle{ruled}{labelfont=normalfont,labelsep=colon,strut=off} % DO NOT CHANGE THIS
\floatstyle{ruled}
\newfloat{listing}{tb}{lst}{}
\floatname{listing}{Listing}
\title{CogRec: A Cognitive Recommender Agent Fusing Large Language Models\\ and Soar for Explainable Recommendation}
\author{
    Jiaxin Hu,
    Tao Wang\thanks{Corresponding author.},
    Bingsan Yang,
    Hongrun Wang
}
\affiliations{
    Sun Yat-Sen University, China\\
    \{hujx57, yangbs3, wanghr56\}@mail2.sysu.edu.cn, wangt339@mail.sysu.edu.cn

}

\usepackage{bibentry}
\begin{document}

\pagestyle{empty}

\maketitle

\begin{abstract}
Large Language Models (LLMs) have demonstrated a remarkable capacity in understanding user preferences for recommendation systems. However, they are constrained by several critical challenges, including their inherent “Black-Box” characteristics, susceptibility to knowledge hallucination, and limited online learning capacity. These factors compromise their trustworthiness and adaptability. Conversely, cognitive architectures such as Soar offer structured and interpretable reasoning processes, yet their knowledge acquisition is notoriously laborious. To address these complementary challenges, we propose a novel cognitive recommender agent called CogRec which synergizes the strengths of LLMs with the Soar cognitive architecture. CogRec leverages Soar as its core symbolic reasoning engine and leverages an LLM for knowledge initialization to populate its working memory with production rules. The agent operates on a Perception-Cognition-Action(PCA) cycle. Upon encountering an impasse, it dynamically queries the LLM to obtain a reasoned solution. This solution is subsequently transformed into a new symbolic production rule via Soar's chunking mechanism, thereby enabling robust online learning. This learning paradigm allows the agent to continuously evolve its knowledge base and furnish highly interpretable rationales for its recommendations. Extensive evaluations conducted on three public datasets demonstrate that CogRec demonstrates significant advantages in recommendation accuracy, explainability, and its efficacy in addressing the long-tail problem.
\end{abstract}


\section{Introduction}

Recommender systems have become a cornerstone of the modern digital information ecosystem, profoundly influencing users' decision-making processes. The recent ascendancy of LLMs \cite{zhao2023survey,wu2024survey} has reinvigorated this field. Leveraging their vast world knowledge acquired from massive text corpora and their powerful contextual understanding, LLMs have demonstrated unprecedented potential in capturing nuanced and dynamic user preferences \cite{geng2022recommendation,cui2022m6}. Techniques such as Chain of Thought (CoT) \cite{wei2022chain} have enabled LLMs to generate seemingly plausible recommendation explanations, marking a significant step towards more transparent systems.

Despite these significant advancements, the current LLM-based recommendation paradigm confronts fundamental challenges: (1)\textbf{“Black-Box” Reasoning}. Although methods like CoT can produce fluent explanatory text, their reasoning process is inherently generative rather than structured. It lacks rigorous logical constraints and state tracking, making it prone to “knowledge hallucination”—generating outputs that are factually incorrect or logically inconsistent, especially in complex, multi-step decision-making scenarios \cite{zhang2025siren,ji2023survey}. (2)\textbf{Limited online learning}. the knowledge within LLMs is embedded in their vast parameters, making real-time, efficient updates from sparse new interaction data difficult~\cite{zheng2025towards,pope2023efficiently}. This renders them ill-suited for adapting to rapidly changing user preferences or addressing the cold-start problem.

In stark contrast, cognitive architectures, exemplified by Soar \cite{laird2019Soar}, offer a distinct alternative. Soar is inherently characterized by several advantages stemming from its production rule-based reasoning engine, explicit goal hierarchy, and its ”chunking” mechanism for learning new rules from experience. It naturally possesses advantages such as high interpretability, strong online learning ability, and efficient symbolic computing. However, Soar's knowledge acquisition bottleneck is pronounced: initial rules must be manually encoded, a process that is both laborious and difficult to scale to complex domains. These complementary challenges motivate the exploration of neuro-symbolic integration~\cite{garcez2023neurosymbolic}. We observe that the semantic understanding and knowledge generation capabilities of LLMs can address Soar's knowledge initialization problem, while Soar's symbolic reasoning and online learning can mitigate the black-box and hallucination issues of LLMs.

To this end, this paper proposes CogRec, a cognitive recommender agent that deeply integrates LLMs with Soar. CogRec designates Soar as its core ”cognitive brain,” responsible for symbolic-level reasoning and learning. It utilizes LLMs as an external knowledge source for initial rule generation and for dynamic querying during an impasse. An innovative Bridge Module facilitates bidirectional neuro-symbolic translation: extracting symbolic rules from the natural language outputs of LLMs, and generating precise prompts from Soar's state to guide the LLMs. This framework operates on a PCA cycle, supports online evolution, and outputs traceable recommendation explanations.The major contributions of this paper are summarized as follows:
\begin{itemize}
    \item We propose CogRec, the first framework to synergize LLMs and the Soar cognitive architecture, delivering an end-to-end solution for explainable recommendation.
    \item We design a neuro-symbolic bridge module for efficient, bidirectional knowledge translation, which enhances system robustness and enables real-time online learning.
    \item Extensive experiments on public datasets validate CogRec's superior performance in accuracy and explainability, along with its effectiveness in mitigating the long-tail issue.
\end{itemize}

\section{Related Work}
\subsection{LLM for Recommendation}

The advent of LLMs has ushered in a paradigm shift in the field of recommender systems~\cite{wu2024survey, zhao2023survey}. Initial research primarily centered on leveraging the powerful semantic representation capabilities of LLMs. Researchers employed LLMs to encode textual descriptions of items or user reviews, thereby augmenting the feature representations of conventional recommender models, such as matrix factorization or graph neural networks \cite{li2023text, li2023preliminary, xu2024sequence}. This approach proved effective in mitigating data sparsity and enhancing performance \cite{wang2023zero}.

As LLMs have grown in scale and capability, the research focus has shifted towards harnessing their generative and reasoning abilities. A prominent line of work has recast the recommendation task as a sequence-to-sequence language generation problem. For instance, P5~\cite{geng2022recommendation} and TALLRec~\cite{bao2023tallrec} introduced a unified pre-training framework that consolidated diverse recommendation sub-tasks—such as sequential recommendation, rating prediction, and explanation generation—under a singular text-to-text paradigm. Such approaches capitalize on the contextual understanding of LLMs, demonstrating significant promise in zero-shot and few-shot recommendation scenarios \cite{li2024ecomgpt,hollmann2022tabpfn}. More recently, researchers have begun to explore the use of LLMs as decision-making agents. These agents are designed to handle complex recommendation tasks by engaging in multi-step reasoning and interacting with external tools, such as search engines and databases~\cite{wang2023recmind, zhao2024recommender}.

\subsection{Cognitive-enhanced Recommendation}

Cognitive-enhanced recommender systems seek to construct models that better align with human decision-making processes by drawing upon theories from cognitive science, thereby enhancing both recommendation accuracy and explainability~\cite{beheshti2020towards}. Early works primarily concentrated on modeling specific cognitive concepts, such as introducing attention mechanisms to simulate user attention allocation~\cite{zhou2018deep} or employing memory networks to emulate users' short- and long-term memory~\cite{cho2023dynamic}. Another line of research has focused on modeling user intent, attempting to disentangle multiple latent intents from user behavior sequences to better comprehend the motivations behind user actions \cite{ma2019learning, li2023preliminary,cen2020controllable}.

The integration of symbolic reasoning into recommendation represents another significant research direction. Several studies have incorporated Knowledge Graphs (KGs) into recommender models, leveraging path-based reasoning over the graph to furnish interpretable rationales for recommendations~\cite{wang2019kgat,li2025knowledge}. For instance, PGPR~\cite{xian2019reinforcement} and CKAN~\cite{wei2023multi} employ reinforcement learning and attention networks to discover inferential paths from users to items within the KG. In contrast to these methods, the impasse-driven online learning paradigm designed for CogRec enables the agent to learn and evolve with greater autonomy and continuity.

\section{Preliminary}
\subsection{Problem Definition}

The problem can be formally defined as follows: Given a set of users, denoted by $\mathcal{U}$, and a set of items, denoted by $\mathcal{I}$, the historical interaction data is represented by a matrix $\mathbf{R}\in\mathbb{R}^{|\mathcal{U}|\times|\mathcal{I}|}$, where $r_{ui}$ denotes the preference of user $u \in \mathcal{U}$ for item $i \in \mathcal{I}$ (e.g., a rating or a click). For a target user $u$, the system's task is to generate a ranked list of recommendations $\mathcal{L}_u = [i_1, i_2, \dots, i_k]$, where each item $i_j \in \mathcal{I}$ is an item not previously interacted with by user $u$.

The cognitive recommendation task proposed in this paper transcends simple item prediction. We aim to construct a cognitive agent whose task is twofold:

\begin{itemize}
\item \textbf{Next-Item Prediction}. To generate a high-quality list of candidate recommendations $\mathcal{L}_u$.
\item \textbf{Generation of Explainable Reasoning Traces}. To produce the symbolic and traceable decision-making process, Eu, that leads to the recommendations.
\end{itemize}

Therefore, the ultimate objective of the CogRec agent is to learn a function $f$ such that: 
\begin{equation}
     f: (\mathcal{U}, \mathcal{I}, \mathbf{R}) \rightarrow (\mathcal{L}_u, \mathcal{E}_u)
\end{equation} 
 
where $\mathcal{E}_u = [e_1, e_2, \dots, e_k]$ is the set of explanations corresponding to the recommendation list $\mathcal{L}_u$. This function, given a user $u$, outputs a tuple $(\mathcal{L}_u, \mathcal{E}_u)$ comprising the recommended items and their corresponding reasoning traces, thereby simultaneously optimizing for both recommendation Accuracy and Transparency.

\subsection{Soar Cognitive Architecture}

Soar is a classic cognitive architecture designed for building general intelligent agents, with its core principle being the simulation of human cognition through symbol-based processes. Its key components are as follows.

\begin{itemize}
\item \textbf{Long-Term Memory (LTM)}: Soar's LTM primarily contains three types of knowledge. Procedural memory is stored as production rules in an IF <conditions> THEN <actions> format. Semantic memory and Episodic memory are used to store factual knowledge and past experiences, respectively.
\item \textbf{Working Memory (WM)}: The WM serves as Soar's dynamic information hub, storing the current task's state, goal hierarchy, and perceptual information. Its content is composed of a collection of Working Memory Elements (WMEs), typically represented in the triplet format of \verb|(object ^attribute value)|. For instance, a user's preference can be represented as \verb|(user <u1> ^likes-genre sci-fi)|.
\item \textbf{Decision Cycle (DC)}: The reasoning process in Soar unfolds through a continuous decision cycle, which comprises five phases: Input, Propose, Select, Apply and Output.
\item \textbf{Impasse and Chunking}: This dual mechanism is central to Soar's learning and problem-solving capabilities and represents a cornerstone of innovation in our CogRec framework.
\end{itemize}

Within our CogRec framework, an impasse serves as the critical trigger for interaction with the LLM. When the agent's endogenous symbolic knowledge is insufficient to make a decision, it suspends its internal reasoning and relays the impasse state information to the LLM to solicit ”external advice.” The chunking mechanism is then responsible for internalizing the LLM-provided solution into a new, generalizable, and permanent production rule, thereby enabling the agent's continuous and autonomous learning and evolution.

\section{Methodology}
\subsection{Framework Overview}
The core tenet of CogRec is to build a recommender agent that emulates human cognitive patterns. Rather than employing an LLM as a monolithic, end-to-end recommender, CogRec designates Soar as its core ”cognitive brain,” tasked with reasoning, decision-making, and learning. The LLM, in turn, serves as a powerful, on-demand external source of knowledge and guidance. As illustrated in Figure 1, the operational workflow of CogRec follows a PCA feedback loop:

\begin{figure*}[t]
\centering
\includegraphics[width=0.9\textwidth]{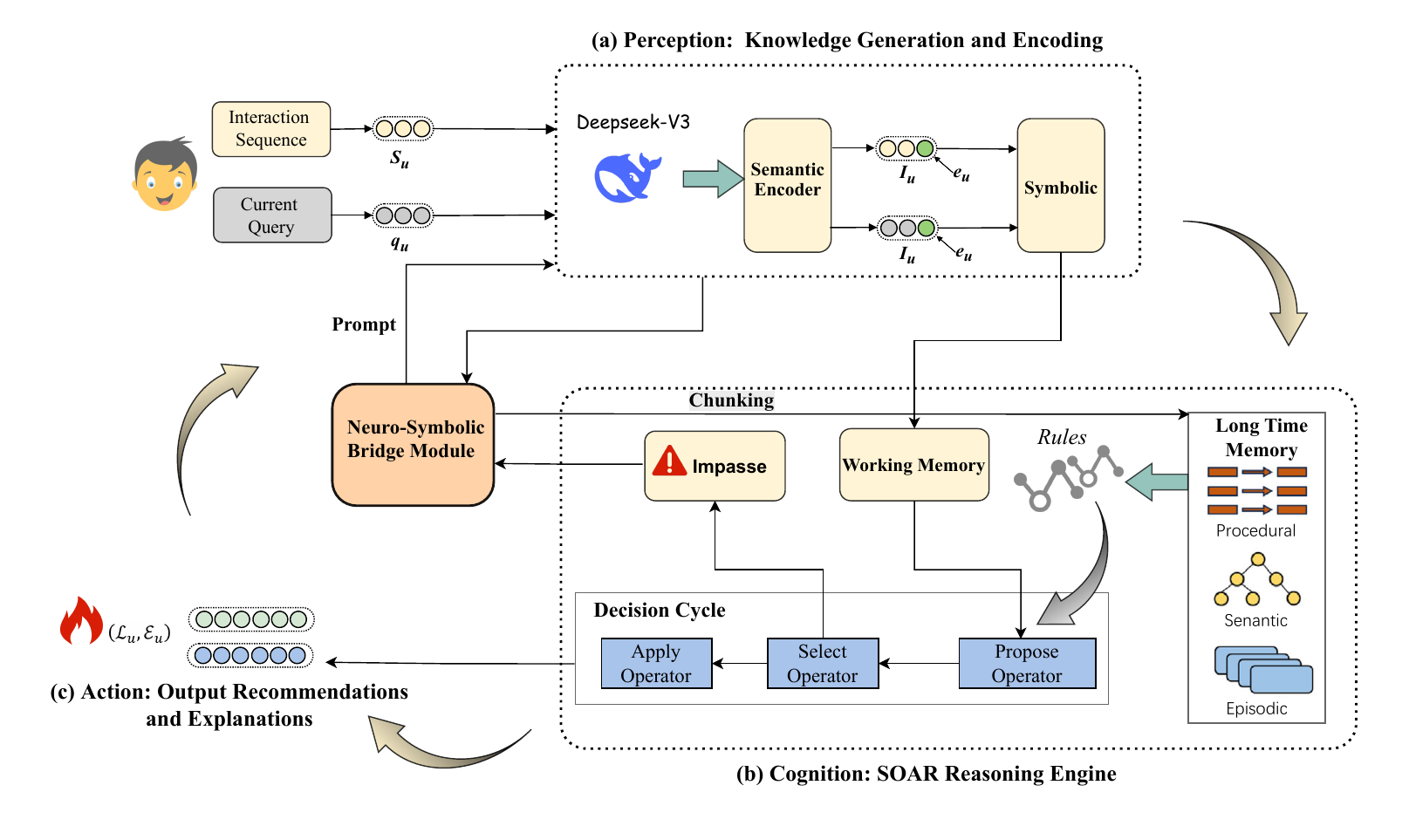} % Reduce the figure size so that it is slightly narrower than the column.
\caption{Overview of the CogRec Framework. The figure illustrates the Soar-centric cognitive cycle, with the LLM serving as an external knowledge module. Neuro-symbolic interaction and learning are facilitated by the Bridge Module. The information flow begins with user input, proceeds through LLM-based encoding into the Soar core for reasoning, and culminates in the output of recommendations and their corresponding explanations.}
\label{fig1}
\end{figure*}

\begin{itemize}
\item \textbf{Perception}. The user interface receives the user's historical interaction sequence $S_u$, and the current query $q_u$. This information is first fed into the LLM module for initial semantic encoding and parsing, which translates the raw input into a symbolic state comprehensible to Soar.
\item \textbf{Cognition}. The Soar core receives this symbolic state and updates its Working Memory. It then leverages the production rules within its procedural memory to reason about the recommendation task, such as decomposing the problem, filtering candidates, and making decisions. If its existing knowledge is sufficient, the process proceeds directly to the action phase. However, if a knowledge gap or conflict is encountered, an impasse is triggered. This is the critical juncture where CogRec's online learning occurs.
\item \textbf{Action \& Learning}. The Soar core ultimately outputs a specific operator, leading to a recommendation. When an impasse occurs, Soar initiates a structured query to the LLM via the bridge module. The LLM's response, after being parsed by the bridge module, is transformed into a new production rule. This new rule is then permanently internalized into Soar's procedural memory through the chunking mechanism. This entire process actualizes the agent's capacity for online learning and evolution.
\end{itemize}

\subsection{Knowledge Generation and Encoding}
We employ the Deepseek-V3 model as the foundational component of our LLM module, where it serves a dual role.

\subsubsection{Initial Semantic Encoder.}
For a given user sequence $S_u$ and query $q_u$, the LLM first transforms them into high-dimensional semantic vectors. Concurrently, it extracts key entities and intents, furnishing a rich contextual foundation for Soar's subsequent symbolic processing. This can be formalized as:

\begin{equation}
    \mathbf{e}_u,I_u=\mathrm{LLM}_\mathrm{encode}(S_u,q_u)
\end{equation} 

where $e_u$ represents the semantic embedding and $I_u$ denotes the extracted preliminary user intent (e.g., “looking for action movies”).

\subsubsection{Knowledge Generator.}
Upon encountering an impasse, Soar queries the LLM, which then generates a potential solution based on the provided context. Leveraging meticulously crafted prompts, we can guide the LLM to generate domain-specific knowledge or inferential steps~\cite{yadav2020distributed}. For instance, during the bootstrapping phase, we inject initial domain knowledge into Soar using the following prompt.

\begin{quote}
\itshape % 将环境内所有文本设为斜体
Prompt: “For a movie recommender system, generate common-sense rules for the ‘sci-fi’ genre. The rules must be in IF-THEN format. For example: IF a user likes ‘Star Wars’, THEN recommend ‘Blade Runner'.”
\end{quote}

\subsection{Cognitive Reasoning Engine}
The cognitive cycle of Soar is precisely mapped onto the recommendation task as follows:

The agent perceives and parses the symbolic information received from the LLM module, which is then used to update its WM. The WM maintains dynamic information pertinent to the current task, including the user model, recommendation goals, and the set of candidate items. For example:

\begin{equation} %\label{eq:wm}
\begin{split}
WM ={}& (\text{goal} < g1 >^{\text{type}} \text{recommend}), \\
          & (\text{user} < u1 >^{\text{history}} < h1 >), \\
          & (\text{candidates} < c1 >^{\text{items}} v1, v2, \dots)
\end{split}
\end{equation}

The Soar decision cycle commences. Production rules $(P_{i}{:}C_{i}\to A_{i})$ stored in procedural memory are matched against the current state of the WM. This involves two substages: (1)\textbf{Proposal}. All rules whose conditions $C_i$ are satisfied by the WM state are activated. Each activated rule proposes an operator $O_i$ to be applied. For example, a rule might propose the operator: \verb|Filter-by-Genre(genre='sci-fi')|. (2)\textbf{Decision}. Based on preferences or learned heuristics, Soar selects the optimal operator, $O^{*}$, from the set of proposed operators.

The selected operator $O^{*}$ is applied, and its resulting actions modify the state of the WM. For instance, after applying the filter operator, the candidate set \verb|<c1>| in the WM would be updated to remove non-sci-fi movies. This cycle continues iteratively until the goal is achieved.

\subsection{Neuro-Symbolic Bridge Module}
This module represents the core innovation of CogRec. It bridges the chasm between the sub-symbolic representations of the LLM and the symbolic representations of Soar, enabling bidirectional information flow and synergy.
\subsubsection{From LLM to Soar.}
When Soar enters an impasse due to insufficient knowledge, the natural language response generated by the LLM must be transformed into symbolic knowledge comprehensible to Soar. This process is performed by the bridge module's Text-to-Chunk Converter, denoted as $\phi_{L}\to S$ ($\phi$ from Language to Symbol).

Consider a scenario where Soar enters an impasse, unable to decide among the candidate set $\{v_A,v_B,v_C\}$, and subsequently queries the LLM. The LLM returns a COT-style response $R_{LLM}$:
\begin{quote}
\itshape % 将环境内所有文本设为斜体
“Recommend ‘Blade Runner 2049’ ($v_A$). Reason (1): The user's history indicates a preference for the cyberpunk genre. Reason (2): ‘Blade Runner 2049’ is a quintessential cyberpunk film.”
\end{quote}

The converter $\phi_{L}\to S$ parses this text into a structured causal pair: Chunkraw = Parse(RLLM) = ⟨{C1, C2}, A1⟩, where the components are:
\begin{equation}
    \mathrm{Chunk}_{\mathrm{raw}}=\mathrm{Parse}(R_{\mathrm{LLM}})=\langle\{C_1,C_2\},A_1\rangle
\end{equation}

where the components are:
\begin{align*} 
    & Condition 1 (C_1): (user <u1>^{preference} {'cyberpunk'}), \\
    & Condition 2 (C_2): (item <v_A>^{genre} {'cyberpunk'}), \\
    & Action (A_1): (propose-operator <op1>^{name} \\
    & select-item^{item} <v_A>)
\end{align*}

Soar's chunking mechanism then ingests this raw chunk and automatically compiles a new and generalizable production rule $P_{new}$, which is then stored in its procedural memory. A simplified representation is:

\begin{lstlisting}[mathescape=true, numbers=none]
$P_{\text{new}}$: sp {
   IF(< s > ^state-is-valid true)
     (< s > ^user < u >)
     (< u > ^preference ?g)
     (< s > ^candidate-item ?i)
     (?i ^genre ?g)
   -->
     (< s > ^operator < o > +)
     (< o > ^name select-item)
     (< o > ^item ?i)
}
\end{lstlisting}

This new rule enables the agent to handle similar situations in the future without re-querying the LLM, thus achieving autonomous knowledge internalization and generalization, which significantly enhances its efficiency and robustness.

\subsubsection{From Soar to LLM.}
To prevent the LLM from generating hallucinations due to vague queries, Soar generates a highly structured prompt when initiating a query. This is accomplished via the bridge module's Symbol-to-Text Converter, $\phi_{S}\to L$. When an impasse occurs, the $WM_{impasse}$ contains a wealth of contextual information. $\phi_{S}\to L$ transforms this context into a precise prompt $Q_{Soar}$, An example of such a prompt is shown in Figure 2.
\begin{equation}
    Q_{\mathrm{Soar}}=\Phi_{\mathrm{S}\to\mathrm{L}}(WM_{\mathrm{impasse}})    
\end{equation}

\begin{figure}[t]
\centering
\includegraphics[width=1.0\columnwidth]{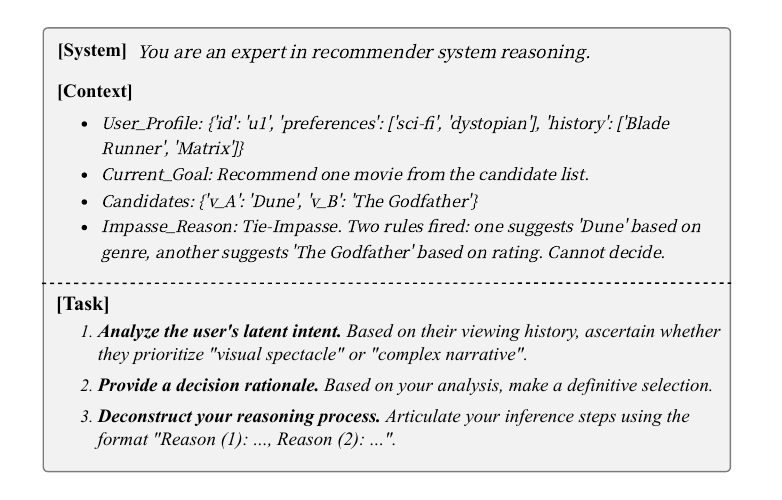} % Reduce the figure size so that it is slightly narrower than the column. Don't use precise values for figure width.This setup will avoid overfull boxes.
\caption{An example of a structured query automatically generated by CogRec's Symbol-to-Text Converter upon encountering a decision impasse.}
\label{fig2}
\end{figure}  

By translating the symbolic state from Soar's working memory (e.g., user profile, goals, candidate set, and the reason for the impasse) into a precise textual prompt, this mechanism effectively guides the LLM to focus on the specific cognitive bottleneck. This, in turn, minimizes knowledge hallucination and ensures the relevance and accuracy of the generated content.

\section{Experiment}
\subsection{Experimental Setup}
\subsubsection{Datasets.}

Our evaluation is performed on three public datasets from distinct domains. These datasets are widely used and are rich in metadata, making them suitable for our study: 1)\textbf{MovieLens-1M (ML-1M)}\cite{harper2015movielens}: A classic dataset for movie recommendation, containing approximately one million ratings from users. We leverage its rich genre information as symbolic attributes for Soar's reasoning processes. 2)\textbf{Amazon Review (Movies)}\cite{mcauley2015image,ni2019justifying}: We select the ”Movies” subset from the Amazon Review dataset. This dataset contains not only ratings but also extensive textual reviews, from which we leverage the LLM to extract user preferences and item attributes. 3)\textbf{Yelp}~\cite{asghar2016yelp}: A dataset comprising user reviews for local businesses. This dataset is selected to evaluate CogRec's generalizability in recommendation scenarios beyond media consumption, such as local services. To ensure data quality, we preprocess all datasets by filtering out users and items with fewer than 10 interactions. Detailed statistics for the datasets are presented in Table 1.

\begin{table}[t]
\centering
%\resizebox{.95\columnwidth}{!}{
\begin{tabular}{lrrrr}
    \toprule % 顶部粗线
        \textbf{Dataset} & \textbf{\#Users} & \textbf{\#Items} & \textbf{\#Inters} & \textbf{\#Sparsity} \\
        \midrule % 中部细线
        ML-1M      & 6,040   & 3,416   & 999,611   & 95.16\% \\
        Movies    & 123,960 & 50,052  & 1,697,533 & 99.97\% \\
        Yelp    & 43,702 & 51,068  & 1,689,188 & 99.93\% \\
        \bottomrule % 底部粗线
\end{tabular}
\caption{Statistics of the experimented datasets.}
\label{table1}
\end{table}

\subsubsection{Evaluation Metrics.}
We employ two standard Top-K ranking metrics to assess recommendation performance: Hit Rate (HR@K) and Normalized Discounted Cumulative Gain (N@K), where K is set to 10 and 20.

\subsubsection{Baselines.}
To comprehensively evaluate our proposed CogRec agent, we select a diverse set of representative models as baselines. These baselines span multiple categories, from conventional methods to state-of-the-art LLM-driven approaches, allowing for a multifaceted performance assessment of CogRec.

\begin{itemize}
\item \textbf{BPR-MF}\cite{rendle2012bpr}: A classic pairwise ranking model based on matrix factorization. It learns latent vectors for users and items by optimizing the probability that a preferred item is ranked higher than an unobserved one, serving as a cornerstone method in collaborative filtering.
\item \textbf{SASRec}\cite{kang2018self}: A powerful sequential recommendation model based on the self-attention mechanism. It predicts the next item by capturing long-range dependencies in the user's interaction history, representing the state-of-the-art in modern sequential models.
\item \textbf{GPT4Rec}\cite{li2023gpt4rec}: An advanced LLM-based method for explainable recommendation. It leverages advanced language models and item content to simultaneously generate personalized recommendations and explainable user interest profiles.
\item \textbf{LLM-Direct}(Deepseek-V3)\cite{liu2024deepseek}: This is a pure LLM-based baseline. We design a meticulous prompt that feeds the user's interaction history and candidate items directly to the Deepseek-V3 model, tasking it with making a direct recommendation.
\item \textbf{Soar}\cite{laird2019Soar}: This is a pure symbolic baseline. We manually encode a small set (approximately 20) of high-quality, generic recommendation rules into the Soar agent but prohibit it from any interaction with the LLM or any online learning.
\end{itemize}

\subsubsection{Implementation Details.}
The CogRec framework is implemented in Python. The Soar core is built upon the pySoar library, while the LLM module interacts with the Deepseek-V3 model via its API. During the knowledge bootstrapping phase, we design approximately 10 high-level prompt templates for each dataset, which are used to generate an initial set of around 500 production rules to populate Soar's procedural memory. For all models, the embedding dimension is set to 128, and they are trained using the Adam optimizer~\cite{loshchilov2017decoupled,kingma2014adam} with a learning rate of 1e-4 and a weight decay of 0.01. Training proceeds for a maximum of 20 epochs. All experiments are conducted on a server equipped with an NVIDIA 4090 GPU.

\subsection{Performance Comparison}
We evaluate CogRec against all baseline models across three datasets, with the results presented in Table 2. Our analysis leads to the following conclusions:
(1) CogRec consistently and significantly outperforms all baseline models across all datasets and metrics. This validates the potent synergy of integrating the rich knowledge from LLMs with the structured reasoning of the Soar architecture.
(2) The performance gains of CogRec over the strong LLM baseline, GPT4Rec, indicate that the symbolic reasoning and online learning mechanisms within our framework can leverage knowledge more effectively than merely generating fluent textual explanations.
(3) CogRec's significant superiority over LLM-Direct provides compelling evidence that the Soar cognitive architecture is far from redundant. It offers crucial capabilities such as goal decomposition, constraint satisfaction, and structured reasoning, which effectively mitigate the logical inconsistencies and knowledge hallucinations that can arise from direct LLM-based decision-making.

\begin{table*}[t]
\centering
\resizebox{\textwidth}{!}{%
\begin{tabular}{l cccc cccc cccc}
    \toprule
    \multirow{2}{*}{\textbf{Models}} & \multicolumn{4}{c}{\textbf{ML-1M}} & \multicolumn{4}{c}{\textbf{Movies}} & \multicolumn{4}{c}{\textbf{Yelp}} \\
    \cmidrule(lr){2-5} \cmidrule(lr){6-9} \cmidrule(lr){10-13}
    & HR@10 & HR@20 & N@10 & N@20 & HR@10 & HR@20 & N@10 & N@20 & HR@10 & HR@20 & N@10 & N@20 \\
    \midrule
    BPR-MF      & 0.1025 & 0.1833 & 0.0512 & 0.0764 & 0.0814 & 0.1521 & 0.0401 & 0.0623 & 0.0622 & 0.1219 & 0.0315 & 0.0501 \\
    SASRec      & 0.2815 & 0.4152 & 0.1743 & 0.2218 & 0.3122 & 0.4589 & 0.2015 & 0.2543 & 0.1988 & 0.3015 & 0.1321 & 0.1764 \\
    GPT4Rec     & \underline{0.2956} & \underline{0.4301} & \underline{0.1899} & \underline{0.2398} & \underline{0.3258} & \underline{0.4712} & \underline{0.2134} & \underline{0.2688} & \underline{0.2105} & \underline{0.3188} & \underline{0.1456} & \underline{0.1892} \\
    LLM-Direct  & 0.2901 & 0.4215 & 0.1832 & 0.2311 & 0.3197 & 0.4655 & 0.2088 & 0.2621 & 0.2053 & 0.3102 & 0.1402 & 0.1823 \\
    Soar        & 0.0511 & 0.0982 & 0.0234 & 0.0356 & 0.0423 & 0.0811 & 0.0198 & 0.0299 & 0.0317 & 0.0624 & 0.0145 & 0.0221 \\
    \midrule
    \textbf{CogRec} & \textbf{0.3124} & \textbf{0.4522} & \textbf{0.2057} & \textbf{0.2581} & \textbf{0.3401} & \textbf{0.4905} & \textbf{0.2269} & \textbf{0.2853} & \textbf{0.2247} & \textbf{0.3351} & \textbf{0.1588} & \textbf{0.2045} \\
    \midrule
    Improve     & 5.7\% & 5.1\% & 8.3\% & 7.6\% & 4.4\% & 4.1\% & 6.3\% & 6.1\% & 6.7\% & 5.1\% & 9.1\% & 8.1\% \\
    \bottomrule
\end{tabular}
}
\caption{Overall performance comparison on the three datasets. The best results are in bold, and the second-best are underlined.}
\label{table2}
\end{table*}

\subsection{Ablation Study}
To validate the necessity of each key component in CogRec, we designed and evaluated the following three ablation variants:

\begin{itemize}
\item \textbf{CogRec w/o LLM-Bootstrap}: This variant removes the LLM-based knowledge bootstrapping phase. Soar begins its learning process from a 'tabula rasa' state, equipped with only five manually coded generic rules.
\item \textbf{CogRec w/o Chunking}: This variant retains the full framework but disables Soar's chunking mechanism after a solution is obtained from the LLM. Consequently, the agent cannot generate new production rules from its experiences.
\item \textbf{CogRec w/o Soar}: This variant is equivalent to the LLM-Direct baseline, removing the entire symbolic reasoning engine.
\end{itemize}

\begin{figure}[t]
\centering
\includegraphics[width=0.9\columnwidth]{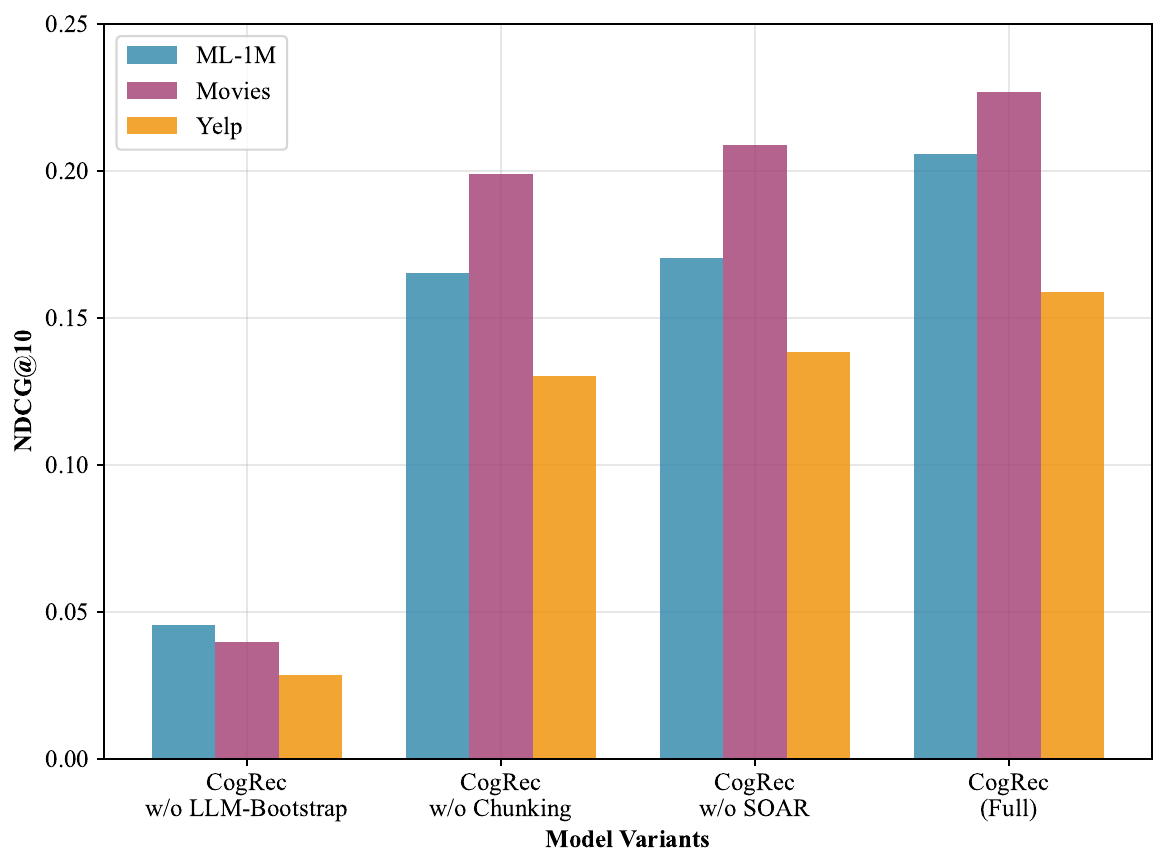} % Reduce the figure size so that it is slightly narrower than the column. Don't use precise values for figure width.This setup will avoid overfull boxes.
\caption{The performance of CogRec and its variants.}
\label{fig3}
\end{figure}  

As Figure \ref{fig3} shows, removing LLM bootstrapping leads to a drastic performance drop, indicating that learning symbolic rules from scratch is exceedingly inefficient. The LLM provides the agent with a high-quality 'innate' knowledge base to start from. After disabling the chunking mechanism, performance degrades significantly, approaching that of LLM-Direct. This demonstrates that online learning is a cornerstone of CogRec's success, enabling it to transform transient solutions into permanent, reusable skills.

\subsection{Further Analysis}
\subsubsection{Learning Curve Analysis.}
As shown in Figure 4, as the number of interactions with users increases, CogRec's LLM-Call Frequency (LCF) declines significantly, eventually converging to a low, stable plateau. This indicates that the agent has learned a sufficient number of rules via chunking to handle most common situations autonomously. In contrast, the LCF of CogRec w/o Chunking remains consistently high, as it is incapable of learning from experience and must re-query the LLM for every similar problem. This provides strong evidence that CogRec’s online learning capability substantially enhances its long-term operational efficiency and autonomy.

\begin{figure}[t]
\centering
\includegraphics[width=0.9\columnwidth]{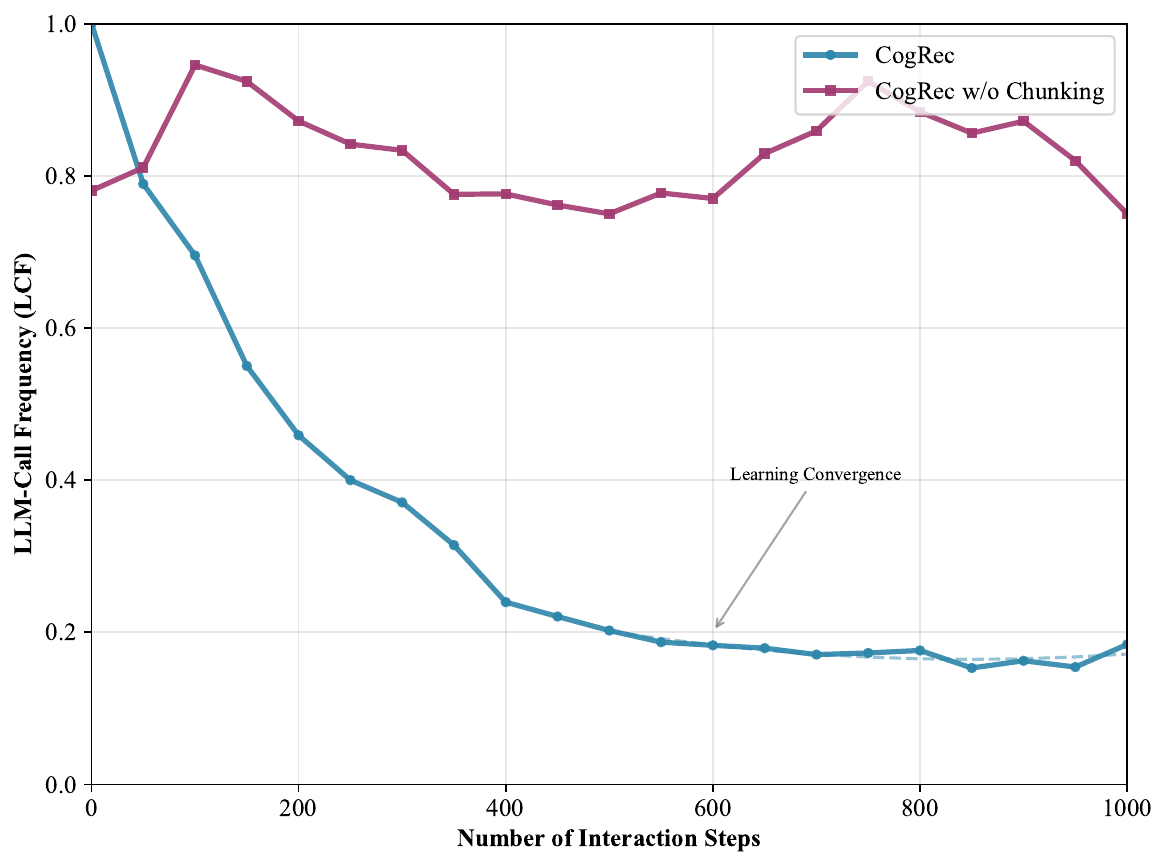} % Reduce the figure size so that it is slightly narrower than the column. Don't use precise values for figure width.This setup will avoid overfull boxes.
\caption{Variation of LCF with the number of interaction steps.}
\label{fig4}
\end{figure} 

\subsubsection{Long-Tail Item Recommendation Performance.}
We partition the items in the test set into head (top 20\%) and long-tail (bottom 80\%) groups based on interaction frequency and evaluate model performance on each. As illustrated in Figure 5, all models experience a performance drop on long-tail items. However, the performance degradation for CogRec is markedly smaller than that of SASRec and LLM-Direct. We attribute this to the fact that CogRec's reasoning does not rely solely on collaborative filtering signals (which are sparse for long-tail items), but rather leverages the common-sense knowledge about item attributes and types provided by the LLM. For instance, even if a niche film has few views, Soar can still make a high-quality recommendation if it knows the film's director or theme matches the user's preferences. This highlights CogRec's robustness in addressing data sparsity issues.

\begin{figure}[t]
\centering
\includegraphics[width=0.9\columnwidth]{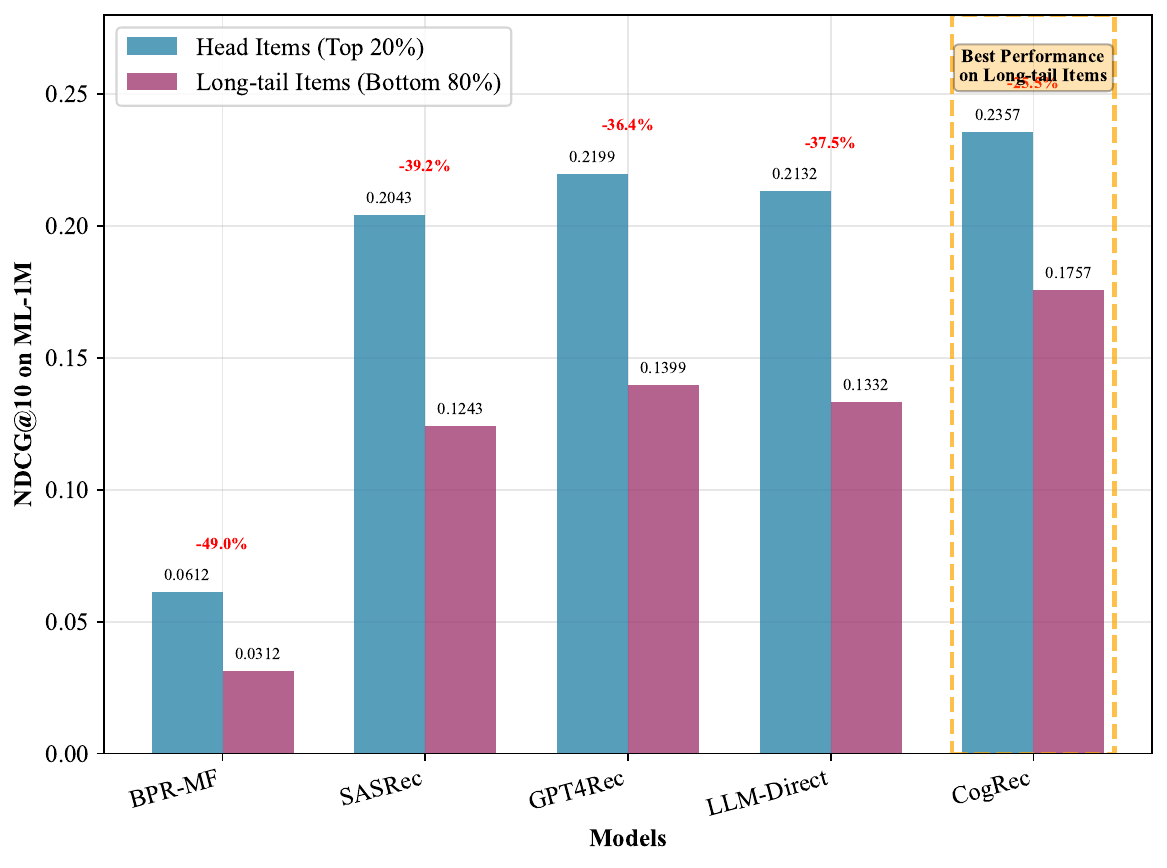} % Reduce the figure size so that it is slightly narrower than the column. Don't use precise values for figure width.This setup will avoid overfull boxes.
\caption{Performance comparison on head and long-tail items (N@10 on ML-1M)}
\label{fig5}
\end{figure} 

\subsection{Case Study}
To intuitively illustrate the explainability of CogRec, we trace the recommendation process for a user with a preference for "sci-fi” and ”mystery” films, as depicted in Figure 6. The process unfolds as follows: (a) the user's partial viewing history is observed, leading to (b) the initialization of Soar's working memory. (c) A 'tie' impasse is then triggered due to multiple viable sci-fi candidates. In response, (d) Soar initiates a structured query to the LLM to probe for deeper-level preferences. (e) The LLM returns a Chain-of-Thought response, identifying a potential user preference for the 'cyberpunk' sub-genre. (f) Subsequently, Soar learns a new production rule via its chunking mechanism, which (g) culminates in the final recommendation of 'Blade Runner 2049,' accompanied by its fully traceable reasoning path.

This case study vividly demonstrates how CogRec reasons through symbolic operations and, when faced with a challenge, resolves it through deep interaction with the LLM to achieve self-evolution. The resulting reasoning trace $\mathcal{T}$, is machine-native and high-fidelity, making it far more trustworthy than any post-hoc generated natural language explanation.

\begin{figure}[t]
\centering
\includegraphics[width=1.0\columnwidth]{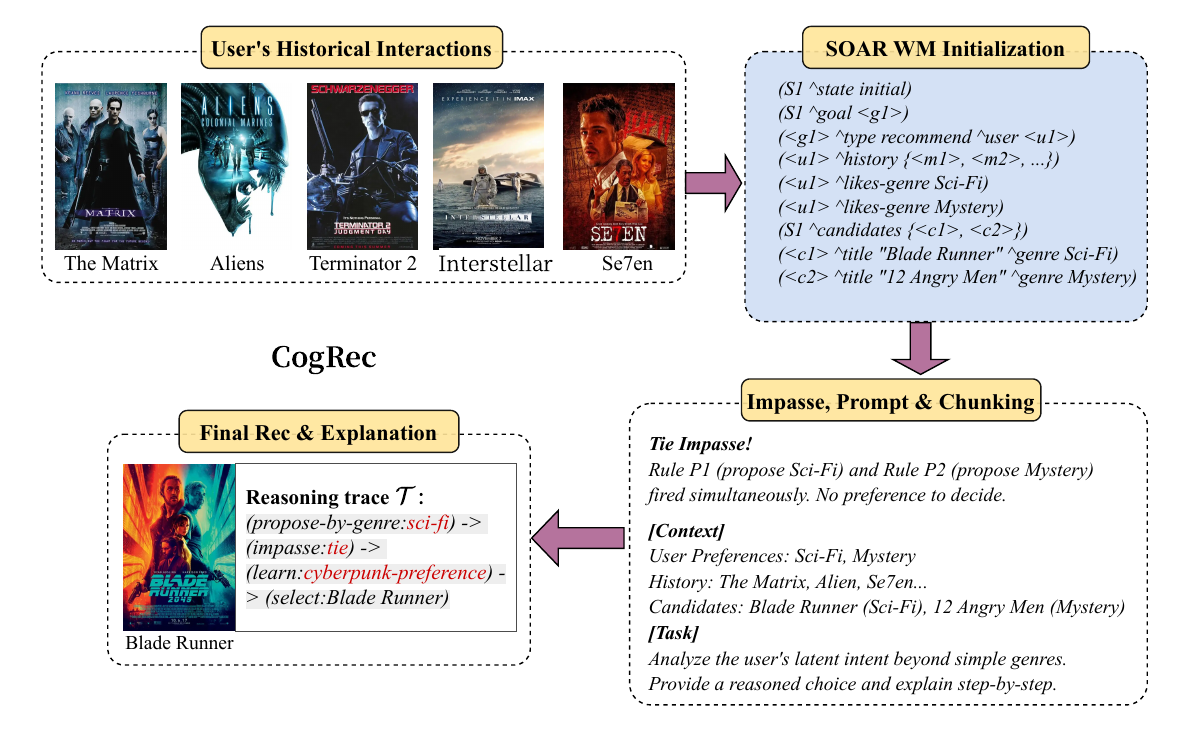} % Reduce the figure size so that it is slightly narrower than the column. Don't use precise values for figure width.This setup will avoid overfull boxes.
\caption{The process of conducting an instance recommendation based on CogRec.}
\label{fig6}
\end{figure} 
\section{Conclusion}
In this paper, we have introduced CogRec, a cognitive recommender agent that tackles the critical challenges of explainability, adaptability, and robustness. It effectively integrates LLMs with the Soar cognitive architecture by employing Soar for symbolic reasoning while utilizing an LLM for knowledge bootstrapping and dynamic impasse resolution. This neuro-symbolic approach establishes a powerful new paradigm, and extensive evaluations have confirmed the efficacy of our proposed CogRec.

%\clearpage

\bibliography{aaai2026}

@article{zhao2023survey,
  title={A survey of large language models},
  author={Zhao, Wayne Xin and Zhou, Kun and Li, Junyi and Tang, Tianyi and Wang, Xiaolei and Hou, Yupeng and Min, Yingqian and Zhang, Beichen and Zhang, Junjie and Dong, Zican and others},
  journal={arXiv preprint arXiv:2303.18223},
  volume={1},
  number={2},
  year={2023}
}

@article{wu2024survey,
  title={A survey on large language models for recommendation},
  author={Wu, Likang and Zheng, Zhi and Qiu, Zhaopeng and Wang, Hao and Gu, Hongchao and Shen, Tingjia and Qin, Chuan and Zhu, Chen and Zhu, Hengshu and Liu, Qi and others},
  journal={World Wide Web},
  volume={27},
  number={5},
  pages={60},
  year={2024},
  publisher={Springer}
}

@inproceedings{geng2022recommendation,
  title={Recommendation as language processing (rlp): A unified pretrain, personalized prompt \& predict paradigm (p5)},
  author={Geng, Shijie and Liu, Shuchang and Fu, Zuohui and Ge, Yingqiang and Zhang, Yongfeng},
  booktitle={Proceedings of the 16th ACM conference on recommender systems},
  pages={299--315},
  year={2022}
}

@article{cui2022m6,
  title={M6-rec: Generative pretrained language models are open-ended recommender systems},
  author={Cui, Zeyu and Ma, Jianxin and Zhou, Chang and Zhou, Jingren and Yang, Hongxia},
  journal={arXiv preprint arXiv:2205.08084},
  year={2022}
}

@article{wei2022chain,
  title={Chain-of-thought prompting elicits reasoning in large language models},
  author={Wei, Jason and Wang, Xuezhi and Schuurmans, Dale and Bosma, Maarten and Xia, Fei and Chi, Ed and Le, Quoc V and Zhou, Denny and others},
  journal={Advances in neural information processing systems},
  volume={35},
  pages={24824--24837},
  year={2022}
}

@article{zhang2025siren,
  title={Siren’s song in the ai ocean: A survey on hallucination in large language models},
  author={Zhang, Yue and Li, Yafu and Cui, Leyang and Cai, Deng and Liu, Lemao and Fu, Tingchen and Huang, Xinting and Zhao, Enbo and Zhang, Yu and Chen, Yulong and others},
  journal={Computational Linguistics},
  pages={1--45},
  year={2025},
  publisher={MIT Press 255 Main Street, 9th Floor, Cambridge, Massachusetts 02142, USA~…}
}

@article{ji2023survey,
  title={Survey of hallucination in natural language generation},
  author={Ji, Ziwei and Lee, Nayeon and Frieske, Rita and Yu, Tiezheng and Su, Dan and Xu, Yan and Ishii, Etsuko and Bang, Ye Jin and Madotto, Andrea and Fung, Pascale},
  journal={ACM computing surveys},
  volume={55},
  number={12},
  pages={1--38},
  year={2023},
  publisher={ACM New York, NY}
}

@article{zheng2025towards,
  title={Towards lifelong learning of large language models: A survey},
  author={Zheng, Junhao and Qiu, Shengjie and Shi, Chengming and Ma, Qianli},
  journal={ACM Computing Surveys},
  volume={57},
  number={8},
  pages={1--35},
  year={2025},
  publisher={ACM New York, NY}
}

@article{pope2023efficiently,
  title={Efficiently scaling transformer inference},
  author={Pope, Reiner and Douglas, Sholto and Chowdhery, Aakanksha and Devlin, Jacob and Bradbury, James and Heek, Jonathan and Xiao, Kefan and Agrawal, Shivani and Dean, Jeff},
  journal={Proceedings of machine learning and systems},
  volume={5},
  pages={606--624},
  year={2023}
}

@book{laird2019soar,
  title={The Soar cognitive architecture},
  author={Laird, John E},
  year={2019},
  publisher={MIT press}
}

@article{garcez2023neurosymbolic,
  title={Neurosymbolic ai: The 3 rd wave},
  author={Garcez, Artur d’Avila and Lamb, Luis C},
  journal={Artificial Intelligence Review},
  volume={56},
  number={11},
  pages={12387--12406},
  year={2023},
  publisher={Springer}
}

@article{zhao2024recommender,
  title={Recommender systems in the era of large language models (llms)},
  author={Zhao, Zihuai and Fan, Wenqi and Li, Jiatong and Liu, Yunqing and Mei, Xiaowei and Wang, Yiqi and Wen, Zhen and Wang, Fei and Zhao, Xiangyu and Tang, Jiliang and others},
  journal={IEEE Transactions on Knowledge and Data Engineering},
  volume={36},
  number={11},
  pages={6889--6907},
  year={2024},
  publisher={IEEE}
}

@inproceedings{li2023text,
  title={Text is all you need: Learning language representations for sequential recommendation},
  author={Li, Jiacheng and Wang, Ming and Li, Jin and Fu, Jinmiao and Shen, Xin and Shang, Jingbo and McAuley, Julian},
  booktitle={Proceedings of the 29th ACM SIGKDD Conference on Knowledge Discovery and Data Mining},
  pages={1258--1267},
  year={2023}
}

@article{li2023preliminary,
  title={A preliminary study of chatgpt on news recommendation: Personalization, provider fairness, fake news},
  author={Li, Xinyi and Zhang, Yongfeng and Malthouse, Edward C},
  journal={arXiv preprint arXiv:2306.10702},
  year={2023}
}

@inproceedings{xu2024sequence,
  title={Sequence-level semantic representation fusion for recommender systems},
  author={Xu, Lanling and Tian, Zhen and Li, Bingqian and Zhang, Junjie and Wang, Daoyuan and Wang, Hongyu and Wang, Jinpeng and Chen, Sheng and Zhao, Wayne Xin},
  booktitle={Proceedings of the 33rd ACM International Conference on Information and Knowledge Management},
  pages={5015--5022},
  year={2024}
}

@article{wang2023zero,
  title={Zero-shot next-item recommendation using large pretrained language models},
  author={Wang, Lei and Lim, Ee-Peng},
  journal={arXiv preprint arXiv:2304.03153},
  year={2023}
}

@inproceedings{bao2023tallrec,
  title={Tallrec: An effective and efficient tuning framework to align large language model with recommendation},
  author={Bao, Keqin and Zhang, Jizhi and Zhang, Yang and Wang, Wenjie and Feng, Fuli and He, Xiangnan},
  booktitle={Proceedings of the 17th ACM conference on recommender systems},
  pages={1007--1014},
  year={2023}
}

@inproceedings{li2024ecomgpt,
  title={Ecomgpt: Instruction-tuning large language models with chain-of-task tasks for e-commerce},
  author={Li, Yangning and Ma, Shirong and Wang, Xiaobin and Huang, Shen and Jiang, Chengyue and Zheng, Hai-Tao and Xie, Pengjun and Huang, Fei and Jiang, Yong},
  booktitle={Proceedings of the AAAI Conference on Artificial Intelligence},
  volume={38(17)},
  pages={18582--18590},
  year={2024}
}

@article{hollmann2022tabpfn,
  title={Tabpfn: A transformer that solves small tabular classification problems in a second},
  author={Hollmann, Noah and M{\"u}ller, Samuel and Eggensperger, Katharina and Hutter, Frank},
  journal={arXiv preprint arXiv:2207.01848},
  year={2022}
}

@article{wang2023recmind,
  title={Recmind: Large language model powered agent for recommendation},
  author={Wang, Yancheng and Jiang, Ziyan and Chen, Zheng and Yang, Fan and Zhou, Yingxue and Cho, Eunah and Fan, Xing and Huang, Xiaojiang and Lu, Yanbin and Yang, Yingzhen},
  journal={arXiv preprint arXiv:2308.14296},
  year={2023}
}

@article{beheshti2020towards,
  title={Towards cognitive recommender systems},
  author={Beheshti, Amin and Yakhchi, Shahpar and Mousaeirad, Salman and Ghafari, Seyed Mohssen and Goluguri, Srinivasa Reddy and Edrisi, Mohammad Amin},
  journal={Algorithms},
  volume={13},
  number={8},
  pages={176},
  year={2020},
  publisher={MDPI}
}

@inproceedings{zhou2018deep,
  title={Deep interest network for click-through rate prediction},
  author={Zhou, Guorui and Zhu, Xiaoqiang and Song, Chenru and Fan, Ying and Zhu, Han and Ma, Xiao and Yan, Yanghui and Jin, Junqi and Li, Han and Gai, Kun},
  booktitle={Proceedings of the 24th ACM SIGKDD international conference on knowledge discovery \& data mining},
  pages={1059--1068},
  year={2018}
}

@inproceedings{cho2023dynamic,
  title={Dynamic multi-behavior sequence modeling for next item recommendation},
  author={Cho, Junsu and Hyun, Dongmin and won Lim, Dong and jae Cheon, Hyeon and Park, Hyoung-iel and Yu, Hwanjo},
  booktitle={Proceedings of the AAAI conference on artificial intelligence},
  volume={37},
  number={4},
  pages={4199--4207},
  year={2023}
}

@article{ma2019learning,
  title={Learning disentangled representations for recommendation},
  author={Ma, Jianxin and Zhou, Chang and Cui, Peng and Yang, Hongxia and Zhu, Wenwu},
  journal={Advances in neural information processing systems},
  volume={32},
  year={2019}
}

@inproceedings{cen2020controllable,
  title={Controllable multi-interest framework for recommendation},
  author={Cen, Yukuo and Zhang, Jianwei and Zou, Xu and Zhou, Chang and Yang, Hongxia and Tang, Jie},
  booktitle={Proceedings of the 26th ACM SIGKDD international conference on knowledge discovery \& data mining},
  pages={2942--2951},
  year={2020}
}

@inproceedings{wang2019kgat,
  title={Kgat: Knowledge graph attention network for recommendation},
  author={Wang, Xiang and He, Xiangnan and Cao, Yixin and Liu, Meng and Chua, Tat-Seng},
  booktitle={Proceedings of the 25th ACM SIGKDD international conference on knowledge discovery \& data mining},
  pages={950--958},
  year={2019}
}

@inproceedings{li2025knowledge,
  title={Knowledge Enhanced Global Graph Contrastive Denoising for Recommendation},
  author={Li, Cheng and Xu, Yong and He, Xin and Zhu, Yujun and Cao, Jinde and Fang, Qun},
  booktitle={International Conference on Intelligent Computing},
  pages={138--149},
  year={2025},
  organization={Springer}
}

@inproceedings{xian2019reinforcement,
  title={Reinforcement knowledge graph reasoning for explainable recommendation},
  author={Xian, Yikun and Fu, Zuohui and Muthukrishnan, Shan and De Melo, Gerard and Zhang, Yongfeng},
  booktitle={Proceedings of the 42nd international ACM SIGIR conference on research and development in information retrieval},
  pages={285--294},
  year={2019}
}

@inproceedings{wei2023multi,
  title={A Multi-Head Attention Network Integrating Knowledge Graph and Collaborative Filtering},
  author={Wei, Yanfei and Miao, Jing and Cheng, Xiaodong and Wang, Yanan},
  booktitle={2023 4th International Symposium on Computer Engineering and Intelligent Communications (ISCEIC)},
  pages={648--654},
  year={2023},
  organization={IEEE}
}

@book{yadav2020distributed,
  title={Distributed artificial intelligence: A modern approach},
  author={Yadav, Satya Prakash and Mahato, Dharmendra Prasad and Linh, Nguyen Thi Dieu},
  year={2020},
  publisher={CRC Press}
}

@article{liu2024deepseek,
  title={Deepseek-v3 technical report},
  author={Liu, Aixin and Feng, Bei and Xue, Bing and Wang, Bingxuan and Wu, Bochao and Lu, Chengda and Zhao, Chenggang and Deng, Chengqi and Zhang, Chenyu and Ruan, Chong and others},
  journal={arXiv preprint arXiv:2412.19437},
  year={2024}
}

@article{harper2015movielens,
  title={The movielens datasets: History and context},
  author={Harper, F Maxwell and Konstan, Joseph A},
  journal={Acm transactions on interactive intelligent systems (tiis)},
  volume={5},
  number={4},
  pages={1--19},
  year={2015},
  publisher={Acm New York, NY, USA}
}

@inproceedings{mcauley2015image,
  title={Image-based recommendations on styles and substitutes},
  author={McAuley, Julian and Targett, Christopher and Shi, Qinfeng and Van Den Hengel, Anton},
  booktitle={Proceedings of the 38th international ACM SIGIR conference on research and development in information retrieval},
  pages={43--52},
  year={2015}
}

@inproceedings{ni2019justifying,
  title={Justifying recommendations using distantly-labeled reviews and fine-grained aspects},
  author={Ni, Jianmo and Li, Jiacheng and McAuley, Julian},
  booktitle={Proceedings of the 2019 conference on empirical methods in natural language processing and the 9th international joint conference on natural language processing (EMNLP-IJCNLP)},
  pages={188--197},
  year={2019}
}

@article{asghar2016yelp,
  title={Yelp dataset challenge: Review rating prediction},
  author={Asghar, Nabiha},
  journal={arXiv preprint arXiv:1605.05362},
  year={2016}
}

@article{rendle2012bpr,
  title={BPR: Bayesian personalized ranking from implicit feedback},
  author={Rendle, Steffen and Freudenthaler, Christoph and Gantner, Zeno and Schmidt-Thieme, Lars},
  journal={arXiv preprint arXiv:1205.2618},
  year={2012}
}

@inproceedings{kang2018self,
  title={Self-attentive sequential recommendation},
  author={Kang, Wang-Cheng and McAuley, Julian},
  booktitle={2018 IEEE international conference on data mining (ICDM)},
  pages={197--206},
  year={2018},
  organization={IEEE}
}

@article{li2023gpt4rec,
  title={GPT4Rec: A generative framework for personalized recommendation and user interests interpretation},
  author={Li, Jinming and Zhang, Wentao and Wang, Tian and Xiong, Guanglei and Lu, Alan and Medioni, Gerard},
  journal={arXiv preprint arXiv:2304.03879},
  year={2023}
}

@article{loshchilov2017decoupled,
  title={Decoupled weight decay regularization},
  author={Loshchilov, Ilya and Hutter, Frank},
  journal={arXiv preprint arXiv:1711.05101},
  year={2017}
}

@article{kingma2014adam,
  title={Adam: A method for stochastic optimization},
  author={Kingma, Diederik P and Ba, Jimmy},
  journal={arXiv preprint arXiv:1412.6980},
  year={2014}
}


\end{document}